\documentclass[runningheads]{llncs}
\usepackage{graphicx}
\usepackage[utf8]{inputenc}
\usepackage{footmisc}
\usepackage{amsmath}
\usepackage{hyperref}
\usepackage{subcaption}
\usepackage{gensymb}
\begin{document}
\title{Can a Robot Shoot an Olympic Recurve Bow? A preliminary study}
%
%
\author{Guilherme Henrique Galelli Christmann \inst{1} \and
Lin Yu-Ren\inst{1} \and
Rodrigo da Silva Guerra\inst{1} \and
Jacky Baltes \inst{1}}
\authorrunning{G. Christmann et al.}
%
\institute{National Taiwan Normal University, Dept. of Electrical Engineering, Taipei, 106, Taiwan \\
\url{wwww.ee.ntnu.edu.tw/ercweb}}
\maketitle              

\begin{abstract}

The field of robotics, and more especially humanoid robotics, has several established competitions with research oriented goals in mind. Challenging the robots in a handful of tasks, these competitions provide a way to gauge the state of the art in robotic design, as well as an indicator for how far we are from reaching human performance. The most notable competitions are RoboCup, which has the long-term goal of competing against a real human team in 2050, and the FIRA HuroCup league, in which humanoid robots have to perform tasks based on actual Olympic events. Having robots compete against humans under the same rules is a challenging goal, and, we believe that it is in the sport of archery that humanoid robots have the most potential to achieve it in the near future. In this work, we perform a first step in this direction. We present a humanoid robot that is capable of gripping, drawing and shooting a recurve bow at a target 10 meters away with considerable accuracy. Additionally, we show that it is also capable of shooting distances of over 50 meters.


\end{abstract}


\section{Introduction}
    \par One of the most admirable traits of modern humans is their continuous endeavor towards greatness. In the physical sense, the Olympics gather the pinnacle of human athletes, performing feats of strength, speed and skill that push the boundaries of human performance. At the same time, in fields such as engineering, researchers design machines and devices that push the limits of creativity and inventiveness. In particular, the field of Humanoid Robotics aims to develop robots that are as capable or even surpass humans in their own tasks and environments.
    \par Looking to bridge the gap between human and robot performance, the scientific community has been organizing research oriented competitions for some time \cite{kitano1995robocup},\cite{simmons19951994}. One of the most prominent examples is the RoboCup Humanoid League with the long-established goal of competing against the winning team of the World Cup at 2050, under official FIFA rules \cite{gerndt2015humanoid}. A closer parallel can be traced to another notable competition: the FIRA HuroCup league, in which humanoid robots perform tasks based on actual Olympic events such as sprint, weightlifting, marathon, basketball, and archery, among others \cite{baltes2017hurocup}.
    \par In the context of having robots competing against humans under the same rules as humans, we believe that archery is the event that has the most potential to achieve this goal in the near future. Archery is a sport that does not require any walking or running, and instead, relies on endurance and strength of the upper body (i.e. a static sport \cite{ertan2003activation}). In addition, accurately hitting a target from some distance requires that the shooting motion be balanced and highly reproducible \cite{soylu2006archery}. By using a robot, reproducibility of the motion is a given to a certain extent, and the main challenge lies in fine control of the upper torso and arms and having enough power to draw and shoot the bow over the required distances. Another important factor to consider is an accurate detection of the target, whose design is usually made up of several concentric circles. This makes it suitable to be detected even with more traditional computer vision techniques, as the one we will present further.
    \par In this paper, we present a humanoid robot that is capable of detecting the target, adjusting the arms position for aiming and reliably drawing and shooting a recurve bow. The rest of this paper is organized as follows: Section \ref{sec:methodology} presents the methodology of the experiments, along with the robot and the recurve bow, Section \ref{sec:results} presents the results from the different experiments and Section \ref{sec:conclusion} concludes the paper and discusses future research paths.
    
\section{Methodology}
    \label{sec:methodology}    

    \par This section will first present the robot and equipment, along with any modifications, followed by the inverse kinematics method to perform the drawing and shooting motions. The measurements for the equipments are presented in the standard units used in the archery domain rather than SI units.
    
    \subsection{Robot and Bow}
        \par
\begin{figure*}[ht]
            \centering
            \includegraphics[width=0.7\textwidth]{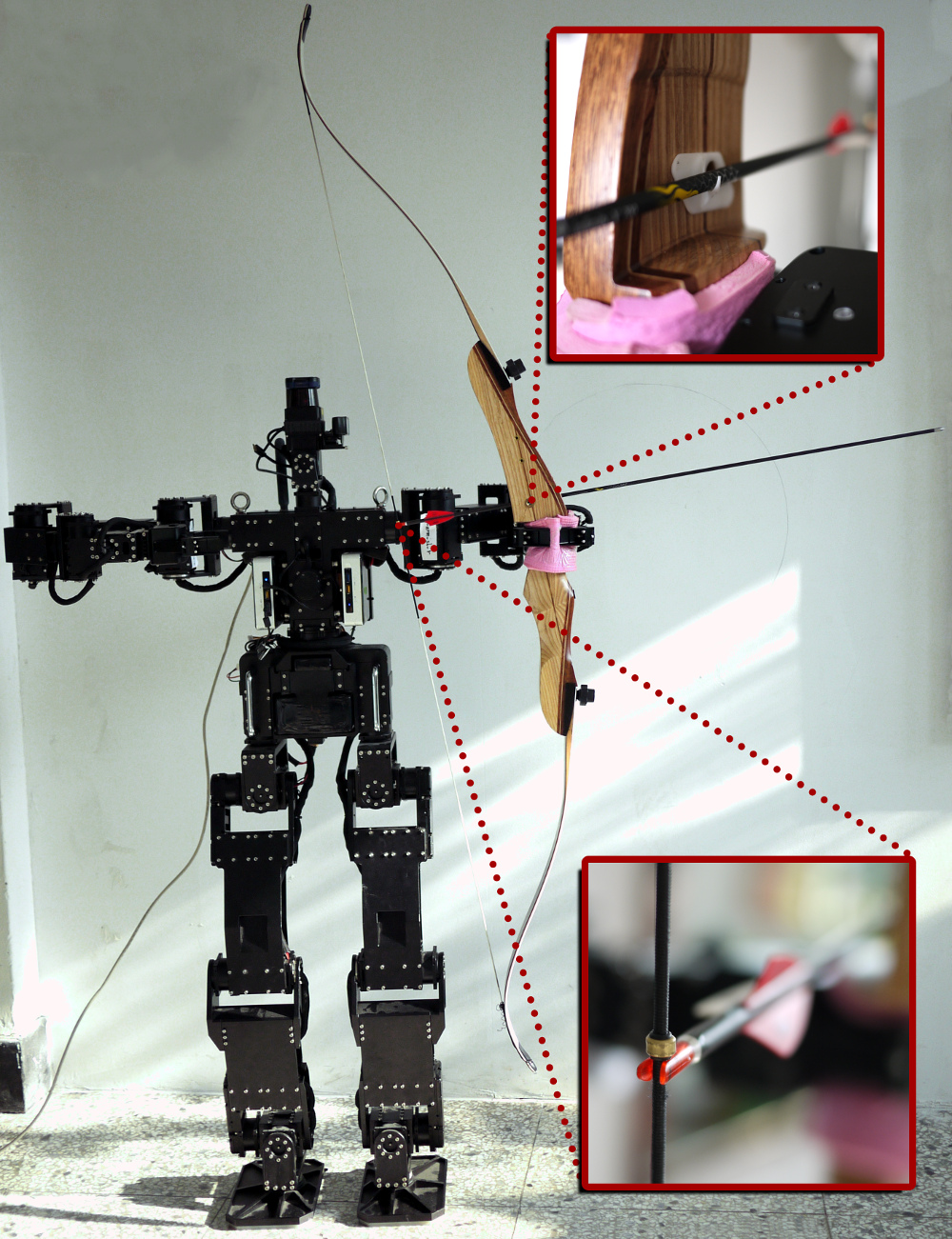}
            \caption{Full picture of the robot THORMANG3 holding the recurve bow with a loaded arrow. The top zoomed picture shows the arrow rest and the bottom picture details the nocking point and how the arrow is attached to the string.}
            \label{fig:thor3_holding_bow}     
        \end{figure*}

 The recurve bow used in this work was a 66 inches long Cartel Sirius Plus\footnotemark\footnotetext{\url{http://www.lancasterarchery.com/cartel-sirius-66-recurve-bow.html}}, shown in Figure \ref{fig:thor3_holding_bow}. This bow is rated with a draw weight of 16 pounds at a draw length of 28 inches. The only modification made to the bow was enveloping the gripping region with ethylene-vinyl acetate (EVA) material. This was needed to make the robot gripper able to hold the bow robustly. We utilized a rubber arrow with a suction cup tip to calibrate the motions and vision of the robot as well as Experiment 1 and 2. For long distance shooting in Experiment 3, to test the limits of the robot's shooting capabilities, we used proper carbon fiber arrows of length 32 inches.
        
        \begin{table}[htb]
            \centering
            \begin{tabular}{|l|l|}
                \hline
                Degrees of Freedom  & 29  \\
                \hline
                Actuator            & 200W x 10 / 100W x 11 / 20W x 8 \\
                \hline
                Computer            & Intel NUC (i5 CPU) -- 8GB RAM DDR4 x2 \\
                \hline
                Camera              & Logitech C920 HD Camera \\
                \hline
                IMU                 & MicroStrain 3DM-GX4-25 \\
                \hline
                Battery             & 22V, 22000mA -- 18.5V, 11000 mA \\
                \hline
                Height              & 137.5cm \\
                \hline
                Weight              & 42 kg \\
                \hline
            \end{tabular}
            \caption{Hardware specifications of THORMANG3.}
            \label{tab:thor3_specs}
        \end{table}
        
        \par The robot used in this work was the humanoid THORMANG3 (Tactical Hazardous Operations Robot 3) from South Korean Manufacturer ROBOTIS. It is a complete bipedal humanoid platform with 29 DoFs, two emdded Intel NUCs, a Full HD camera and several other sensors. The hardware specifications of THORMANG3 are presented in Table \ref{tab:thor3_specs}. A full picture of the robot holding the bow is presented in Figure \ref{fig:thor3_holding_bow}. The zoomed picture at the top shows the arrow rest in detail and the one at the bottom details how the arrow is attached to the string, where the metal ring is called the nock.
        
    \subsection{Vision System}
    
    \par The target is detected using an approach based on finding three concentric circles with the Hough Transform. The vision system performs the following sequence of operations:
    
    \begin{enumerate}
        \item Convert the image to grayscale.
        \item Smooth the image applying a Gaussian Blur with a kernel of size 5.
        \item Detect circles in the image using the Hough Transform. This step requires some tuning, depending on the size of the circles on the target and the distance of the target to the robot. For Experiment 1, with the target 10 meters away, the minimum radius was 10 pixels and the maximum radius was 30 pixels, and the Hough accumulator with same resolution as the image.
        \item For each detected circle, we compute a vector of the pixel-by-pixel differences in eight different directions, starting from the center of the circle. The directions are up, down, left, right, as well as the four diagonals. Pixel-by-pixel differences smaller than a threshold of 40 are set to 0, in order to ignore small changes in pixel values.
        \item We perform non-maximum suppression on the difference vectors, using a range of $\pm10\%$ of the length of the difference vector.        
        \item For each of the difference vectors, the indices of the first, second and third local maxima are found, labeling the difference vector as a positive detection. Since we're looking for three concentric circles, if three local maxima can't be found, the difference vector is labeled as a negative detection.
        \item We estimate the increase of the radius of the circles $r_{\text{avg}}$, by assuming that the first circle has a radius of $r_1 = r_{\text{avg}}$, the second circle $r_2 = 2 \cdot r_{\text{avg}}$, and the third circle $r_3 = 3 \cdot r_{\text{avg}}$. Then, we check that each detected radius falls within a tolerance of 50\% of the predicted radius $r_{\text{avg}}$.
        \item If at least 5 difference vectors are labeled as a positive detection, the location of the target is defined as the center of the detected circles.

    \end{enumerate}
        
    \subsection{Shooting Method}
        \par The shooting can be decomposed in two steps. First, we determine the location of the target in the camera image using a Hough Transform \cite{ballard1981generalizing} approach to detect three concentric circles. This is done in order to effectively aim at the target. Secondly, for the actual draw and release motion, we independently control each arm using inverse kinematics, specifically by computing the pseudoinverse of the Jacobian matrix \cite{craig2009introduction}.
        \par
\begin{figure}[htb]
            \centering
            \includegraphics[width=0.48\textwidth]{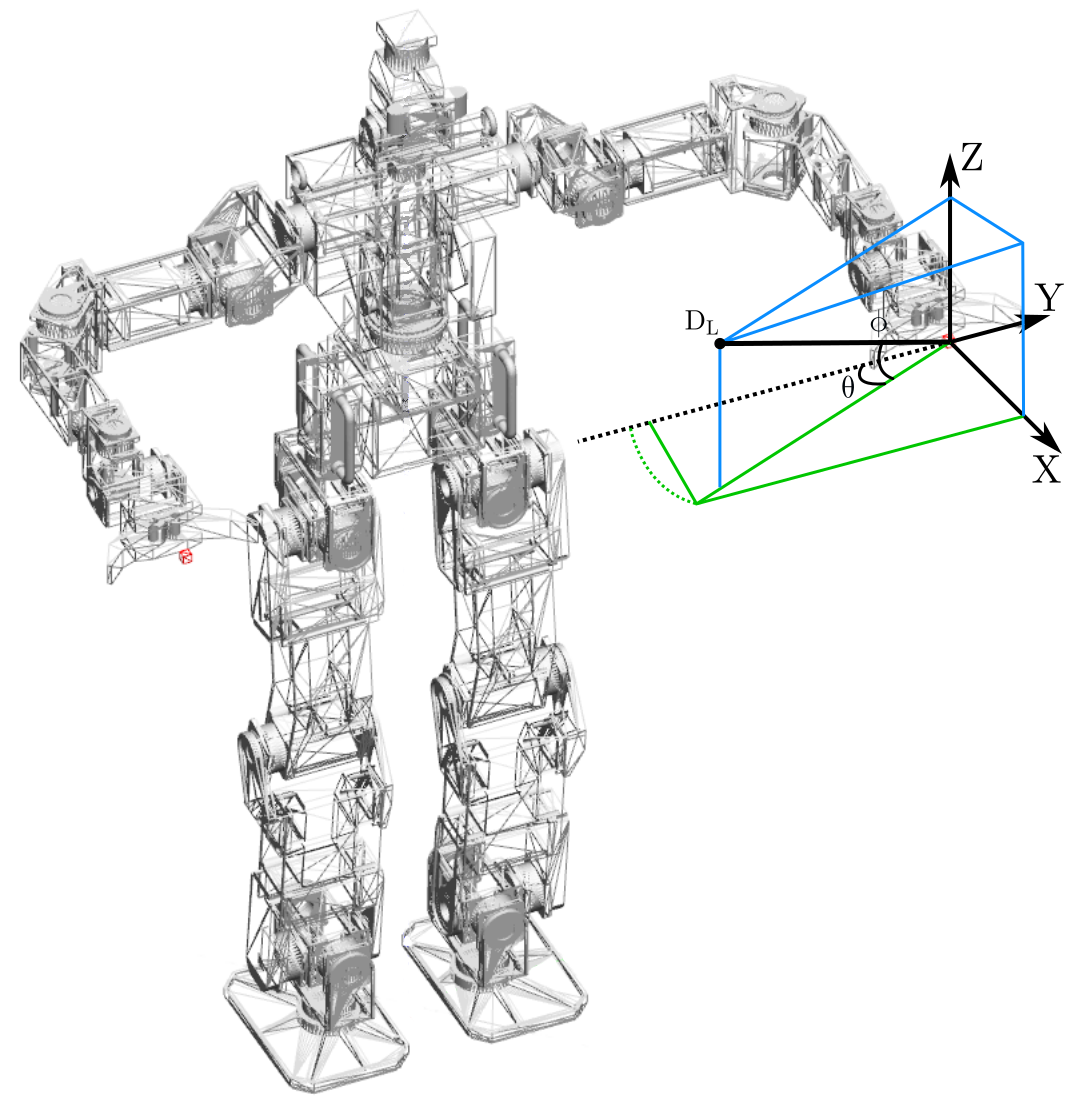}
            \caption{The reference frame is fixed on the left gripper of the robot. The draw vector has length $D_L$ and same origin as the reference frame.}
            \label{fig:ref_frame}
        \end{figure}

 The left gripper of the robot is used to hold the bow, while the right gripper is used to draw and release the string, as shown in Figure \ref{fig:thor3_holding_bow}. The right gripper has an offset of $-\pi/2$ radians in Yaw compared to the left gripper. To guarantee a good release it is important that the drawing motion is performed along the vector from the arrow rest to the nock. This is called the draw vector. Deviations from the draw vector when performing the drawing and releasing motions can cause the arrow to shoot off-target or to be knocked off the bow. The following set of equations demonstrate how to compute any point along the draw vector:

        \begin{subequations}
            \label{eq:deltas}
            \begin{align}
                \Delta X &= sin(\theta) \cdot D_L \\
                \Delta Y &= cos(\theta) \cdot cos(\phi) \cdot D_L \\
                \Delta Z &= sin(-\phi) \cdot D_L \\
                R_x &= L_x + \Delta X \\ 
                R_y &= L_y + \Delta Y \\
                R_z &= L_z + \Delta Z
            \end{align}
        \end{subequations}
        
        \noindent where $\theta$ and $\phi$ are the Yaw and Roll for the left gripper respectively, and, in that order, $D_L$ is the distance along the draw vector, starting from the left gripper, $L_x, L_y, L_z$ are the coordinates of the left gripper, computed through forward kinematics, and $R_x$, $R_y$ and $R_z$ are the coordinates of the right gripper. The angles and positions are computed with the reference frame fixed on the left gripper, as shown on Figure \ref{fig:ref_frame}. Yaw ($\theta$) is defined as the rotation along the $Z$ axis (green) and Roll ($\phi$) is the rotation around the $X$ axis (blue).
        
    \par The whole act of shooting the bow is then realized in the following steps:
        \begin{enumerate}
            \item The arrow is assumed to be correctly placed and nocked on the bow by a human, as shown in Figure \ref{fig:thor3_holding_bow}.
            \item After detecting the target, the orientation (Yaw and Roll) of the left gripper is adjusted to correctly aim at the target.
            \item The string rests at 22 cm from the arrow rest. Plugging this value in $D_L$ from Equations \ref{eq:deltas}, the right gripper is moved on top of the string.
            \item Right gripper is closed.
            \item With the gripper closed, we can plug a new value in $D_L$ for the desired draw length and slowly draw the bow to the computed point along the draw vector.
            \item Open the right gripper, releasing the arrow at the aimed direction with force proportional to the draw length.
        \end{enumerate}
        
    \begin{figure}[h]
        \centering
        \begin{subfigure}[t]{0.48\textwidth}
            \centering
            \includegraphics[height=5.85cm]{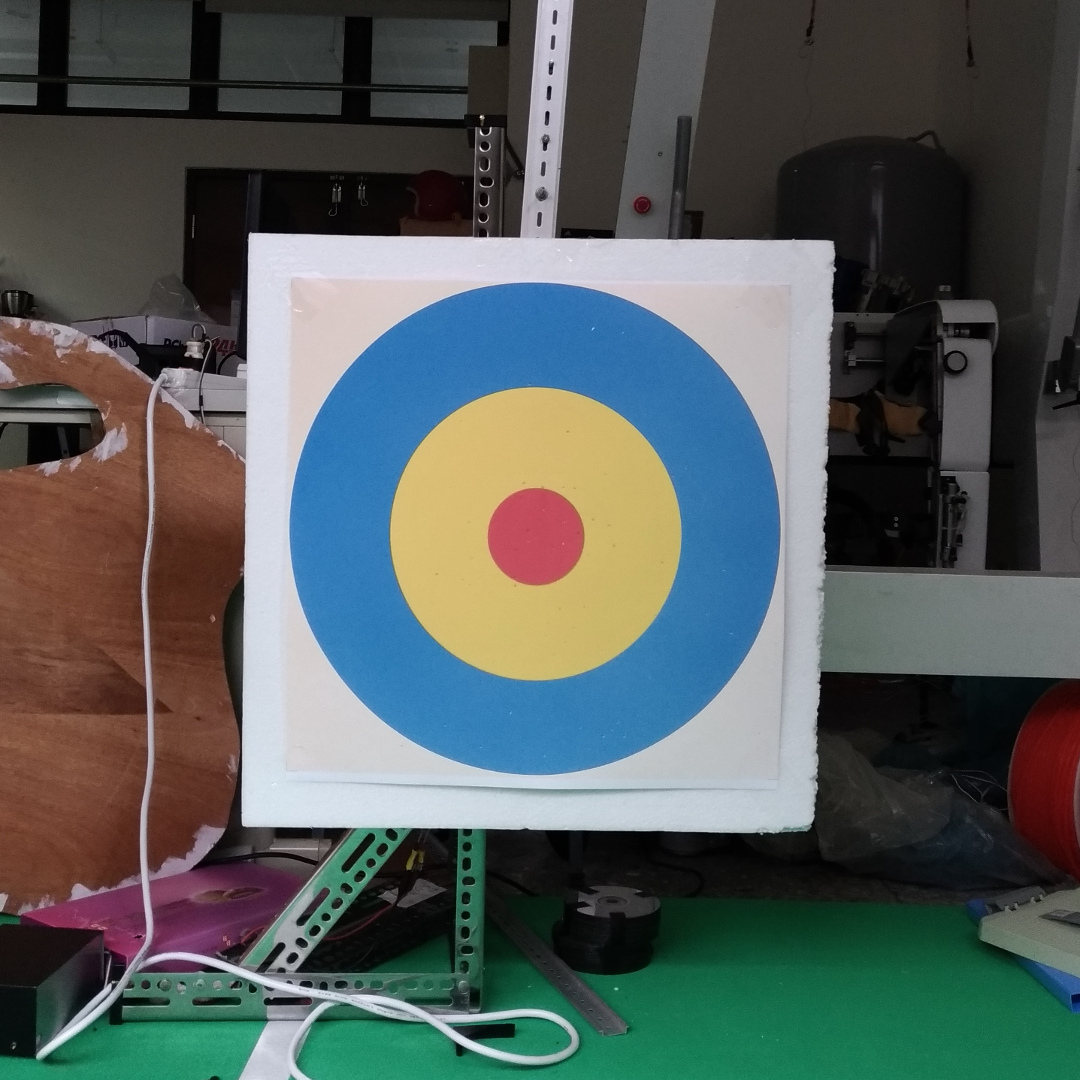}
            \caption{Original image.}
            \label{fig:target1}
        \end{subfigure}
        ~
        \begin{subfigure}[t]{0.48\textwidth}
            \centering
            \includegraphics[height=5.85cm]{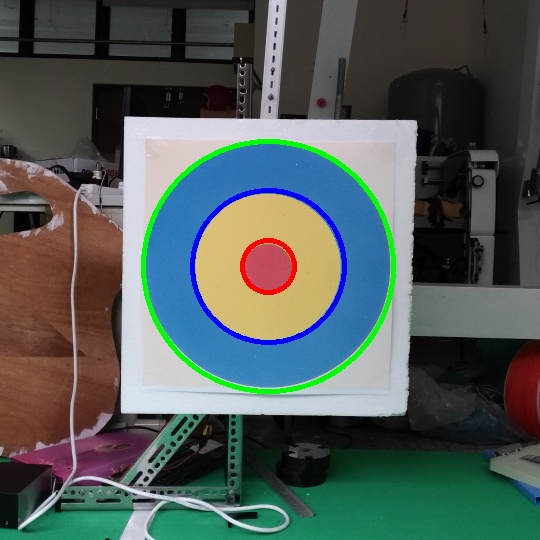}
            \caption{Detected target.}
            \label{fig:target2}
        \end{subfigure}
        \caption{Target used for Experiments 1 and 2 and its detection result using the Hough Transform approach.}
    \end{figure}              
        
    \par The second step requires some calibration in order to determine the correct adjustments to the orientation of the shot. Since one of our main focus in this study is the repeatability of the motion, we opted for a simple proportional adjustment for the Yaw and used a fixed value for Roll. It should be noted that Yaw affects the shot sideways while Roll affects upwards. To calibrate, we made some shots with Yaw value of 0\degree and placed the target in center of the spread of the shots. Then, the $X$ coordinate where the target is detected in the image is defined as the reference $X_\text{ref}$. Next, the target is moved sideways and we perform some shots by manually inputting the Yaw values, searching for the value that correctly hits the target. The proportional gain is then given by: 
    \begin{equation}
            K_p = \frac{Yaw}{X - X_\text{ref}}
    \end{equation}

\section{Results}
    \label{sec:results} 

    \par To evaluate the capabilities of the robot shooting the bow we performed three experiments.
    \begin{enumerate}
        \item Shooting at a wall 10 m away from the feet of the robot, without applying any orientation adjustments, i.e. $Yaw=0\degree$ and $Roll=0\degree$. The draw length $D_L$ used was 65 cm and for safety reasons a rubber arrow was used. A total of 10 shots were performed and the robot was not moved or replaced between the shots.
        \item Shooting at a target 10m away from the feet of the robot using vision for target detection and orientation adjustments. The target is a circle of 49 cm of diameter, hanged with its center at 114.5cm from the ground. The innermost circle has a diameter of 9.7 cm. A picture of the target in presented in Figure \ref{fig:target1}. The draw length $D_L$ used was 65 cm and the same rubber arrow from the previous experiment was used. A total of 10 shots were performed and the robot was not moved or replaced between the shots.
        \item Shooting long distances using a fiber glass arrow in an open field. Different draw lengths $D_L$ and $Roll$ values were utilized to determine the maximum range of the shooting. Since the goal of this experiment was simply to determine the maximum shooting distance, no vision system or target detection was employed. 
    \end{enumerate}
    
    \par It should be noted that at the time of the realization of Experiment 1, the target was already attached to the wall, and Experiment 2 was performed right after finishing the first experiment, without moving or readjusting the robot's position.

    \par
\begin{figure}[h]
        \centering
        \begin{subfigure}[t]{0.48\textwidth}
            \centering
            \includegraphics[height=5.85cm]{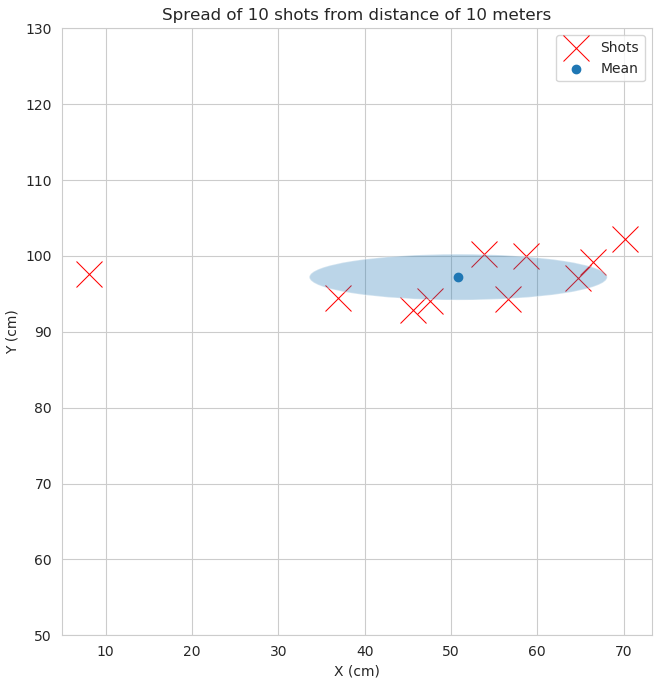}
            \caption{Experiment 1: shooting 10 arrows at a wall without orientation adjustments.}
            \label{fig:shots_spread}
        \end{subfigure}
        ~
        \begin{subfigure}[t]{0.48\textwidth}
            \centering
            \includegraphics[height=5.85cm]{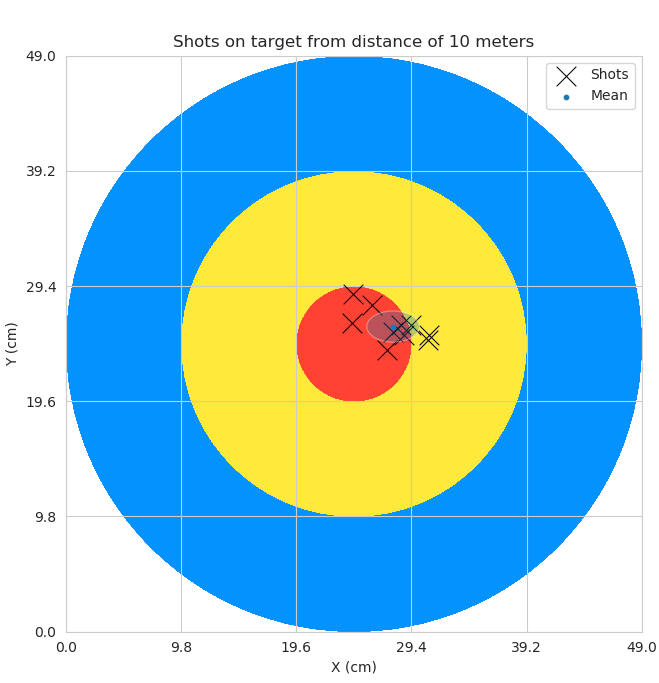}
            \caption{Experiment 2: shooting 10 arrows at the target 10m away.}
            \label{fig:shots_on_target}
        \end{subfigure}
        \caption{Results for Experiments 1 and 2.}
    \end{figure}

 Figure \ref{fig:shots_spread} shows the spread of the 10 shots from Experiment 1 against the wall. The $Y$ axis is measured from the ground. The largest distance between any 2 shots was 62.14 cm in the $X$ axis and 9.34 cm in the $Y$ axis. Furthermore, the variation was 297.75 cm for the $X$ axis and 9.14 cm for the $Y$ axis. During this experiment it became apparent that the robot would wobble a bit after each shot and rotate slightly to the right. This can be noticed on the spread of the arrows shown.
    \par Figure \ref{fig:shots_on_target} shows the 10 shots at the target from Experiment 2. The $X$ and $Y$ axis are relative to the target, and the units are displayed in centimeters. The mean $X$ and $Y$ shot is located at 27.85 cm and 25.96 cm respectively. The largest distance was 6.6 cm in the $X$ axis and 4.7 cm in $Y$ axis. The variances were 5.01 cm and 1.74 cm for the $X$ and $Y$ axis respectively.
    \par
\begin{figure}
        \centering
        \includegraphics[width=0.65\textwidth]{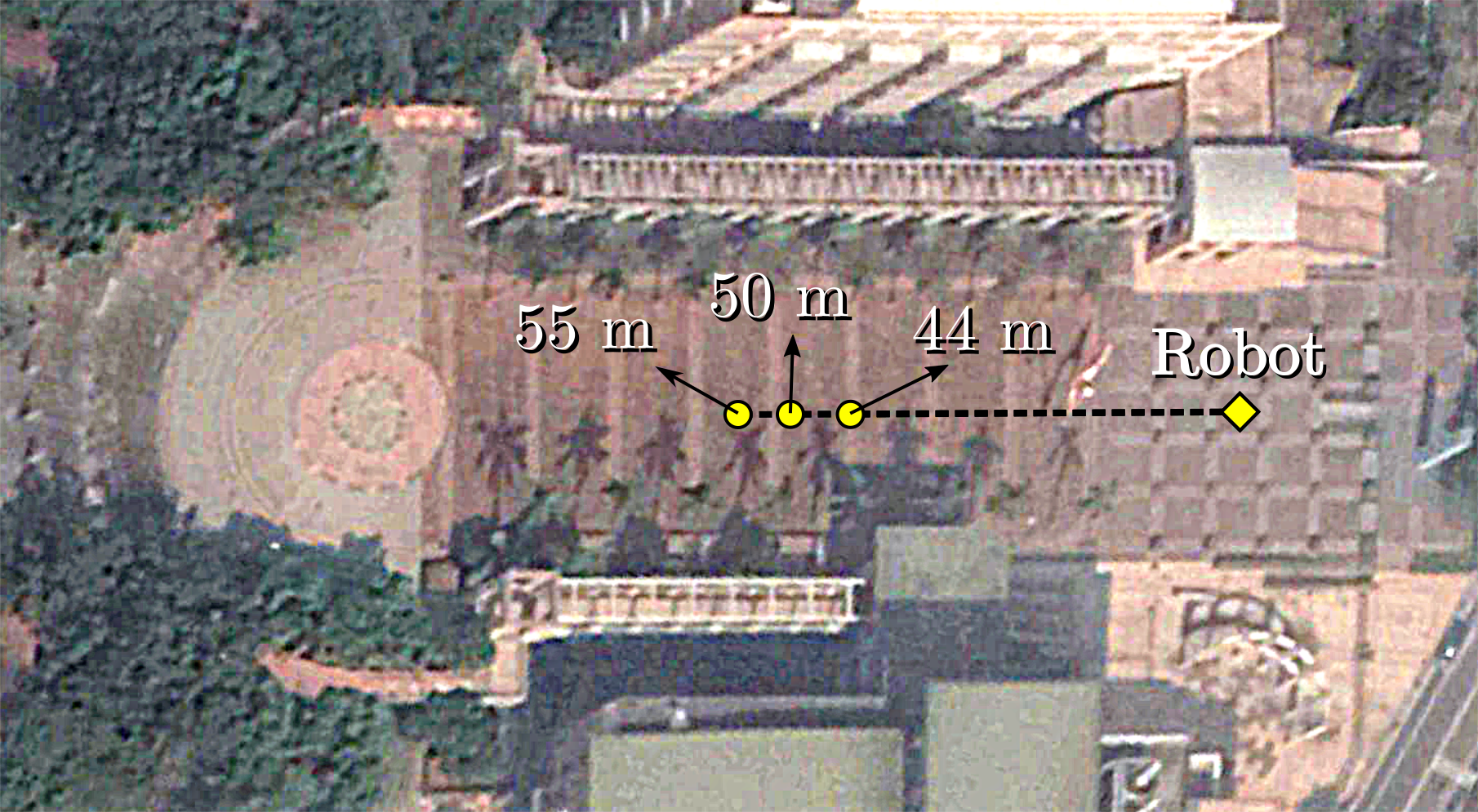}
        \caption{Satellite image of the open field where Experiment 3 was realized, as well as a depicted approximation of where the arrows landed and their distance.}
        \label{fig:arrows_dist}
    \end{figure}

 The results of Experiment 3, where the intent was determining the maximum shooting distance, are presented in Table \ref{tab:shot_dists}. It shows the distance of the shot arrows for different draw distances $D_L$ and $Roll$ values. The distances were measured by recording a video of each shot and observing where the arrows landed. After, we used Google Earth\footnote{https://www.google.com/earth/} to measure the approximate distance for each arrow, depicted in Figure \ref{fig:arrows_dist}.
    
    \begin{table}[h]
        \centering
        \begin{tabular}{|c|c|c|}
            \hline
            \textbf{$D_L$ (cm)} & \textbf{$Roll$} & \textbf{Shot Distance (m)} \\
            \hline
            70 & 4\degree & 44 \\
            70 & 10\degree & \textbf{55} \\
            65 & 12\degree & 50 \\
            \hline
            
        \end{tabular}
        \caption{Results of Experiment 3, distance of shot for different draw lengths and $Roll$ values.}
        \label{tab:shot_dists}
    \end{table}

\section{Conclusion and Future Work}
    \label{sec:conclusion}  

    \par This work presented a humanoid robot that is capable of holding, drawing and shooting a recurve bow at a target 10 meters away with considerable accuracy. In addition, the robot was also capable of shooting distances of over 50 meters. Real world competitions are realized with the target at a distance of 18 m indoors or outdoors with distances of 60 m or 70 m. To the best of our knowledge, this is the first time that a humanoid robot has shown such potential to compete in a real-world archery environment under the same rules as humans.
    \par In the future, we plan to take the robot to an indoor shooting range and use fiber glass arrows to measure its scoring capabilities in comparison to human athletes. We also plan to experiment with bows of heavier draw weight, which would allow us to shoot longer distances and possibly hit targets 60 m and 70 m away.
    
\section{Acknowledgements}
    \par This work was financially supported by the ``Chinese Language and Technology Center'' of National Taiwan Normal University (NTNU) from The Featured Areas Research Center Program within the framework of the Higher Education Sprout Project by the Ministry of Education (MOE) in Taiwan, and Ministry of Science and Technology, Taiwan, under Grants No. MOST 108-2634-F-003-002, MOST 108-2634-F-003-003, and MOST 108-2634-F-003-004 (administered through Pervasive Artificial Intelligence Research (PAIR) Labs), as well as MOST 107-2811-E-003-503. We are grateful to the National Center for High-performance Computing for computer time and facilities to conduct this research.

%
%
\bibliographystyle{splncs04}
\bibliography{bibliography}

\end{document}